\documentclass{article}
\usepackage{spconf,amsmath,graphicx}
\usepackage{multirow}
\usepackage{caption}
\let\url\relax

\usepackage{enumitem}
\setlist{nosep, leftmargin=14pt}

\usepackage{mwe} 

\title{Lost in Distortion: Uncovering the Domain Gap Between Computer Vision and Brain Imaging - A Study on Pretraining for Age Prediction}
\name{Yanteng Zhang$^{1,*}$, 
Songheng Li$^{2}$, 
Zeyu Shen$^{3}$, Qizhen Lan$^{4}$, Lipei Zhang$^{5}$, 
Yang Liu$^{6}$,
Vince Calhoun$^{1}$\thanks{*Corresponding author: yntn32@outlook.com}%
  \thanks{Thanks to NIH RF1AG063153 funding.}}

\address{
$^{1}$Center for Translational Research in Neuroimaging and Data Science (GSU, GATech, Emory), USA; \\
$^{2}$College of Artificial Intelligence, Chengdu University of Information Technology, Chengdu, China; \\
$^{3}$Department of Applied Mathematics and Statistics, Johns Hopkins University, Baltimore, USA; \\
$^{4}$McWilliams School of Biomedical Informatics, University of Texas Health Center at Houston, USA; \\
$^{5}$Department of Applied Mathematics and Theoretical Physics, University of Cambridge, UK; \\
$^{6}$School of Biomedical Engineering and Imaging Sciences, King's College London, UK
}

\begin{document}
%
\maketitle
\begin{abstract}
Large-scale brain imaging datasets provide unprecedented opportunities for developing domain foundation models through pretraining. However, unlike natural image datasets in computer vision, these neuroimaging data often exhibit high heterogeneity in quality, ranging from well-structured scans to severely distorted or incomplete brain volumes. This raises a fundamental question: can noise or low-quality scans contribute meaningfully to pretraining, or do they instead hinder model learning? In this study, we systematically explore the role of data quality level in pretraining and its impact on downstream tasks. Specifically, we perform pretraining on datasets with different quality levels and perform fine-tuning for brain age prediction on external cohorts. Our results show significant performance differences across quality levels, revealing both opportunities and limitations. We further discuss the gap between computer vision practices and clinical neuroimaging standards, emphasizing the necessity of domain-aware curation to ensure trusted and generalizable domain-specific foundation models.
\end{abstract}
\section{Introduction}
\label{sec:intro}

The advent of foundation models has profoundly reshaped machine learning, with large-scale pretraining driving major breakthroughs in natural image understanding. Frameworks such as MAE~\cite{he2022masked}, DINO~\cite{simeoni2025dinov3}, and SimCLR~\cite{chen2020simple} have demonstrated that general purpose visual representations can be learned from massive unlabeled datasets, leading to strong transferability and data efficiency in a variety of downstream tasks~\cite{xiao2025medicalimages}. Inspired by this paradigm, the medical imaging community has begun to explore self-supervised pretraining in modalities such as MRI and CT to reduce dependence on annotated data and enhance model generalization for applications including disease diagnosis, brain age prediction, and structural abnormality detection~\cite{gu2025vision}\cite{joo2023brain}. 

However, directly transferring such strategies from natural image analysis to neuroimaging presents unique challenges. Unlike natural image datasets that are typically clean and semantically diverse, neuroimaging repositories often contain scans with artifacts, intensity distortions, or even missing brain regions caused by acquisition variability across scanners, sites, and protocols. This inherent heterogeneity raises a critical question: does incorporating low-quality scans enrich pretraining by increasing data diversity, or does it simply introduce noise that undermines downstream generalization? Addressing this question is essential for understanding whether scaling up heterogeneous neuroimaging data truly benefits self-supervised learning or compromises representation consistency. In addition, brain imaging exhibits considerable complexity due to differences in imaging mechanisms, anatomical constraints, and multi-center acquisition settings~\cite{guan2021multi}\cite{dartora2024deep}. The non-standardization and heterogeneity introduced by various scanners and preprocessing pipelines make it challenging to build stable and transferable self-supervised representations.

\begin{figure*}[!t]
\centering
\includegraphics[width=0.95\linewidth]{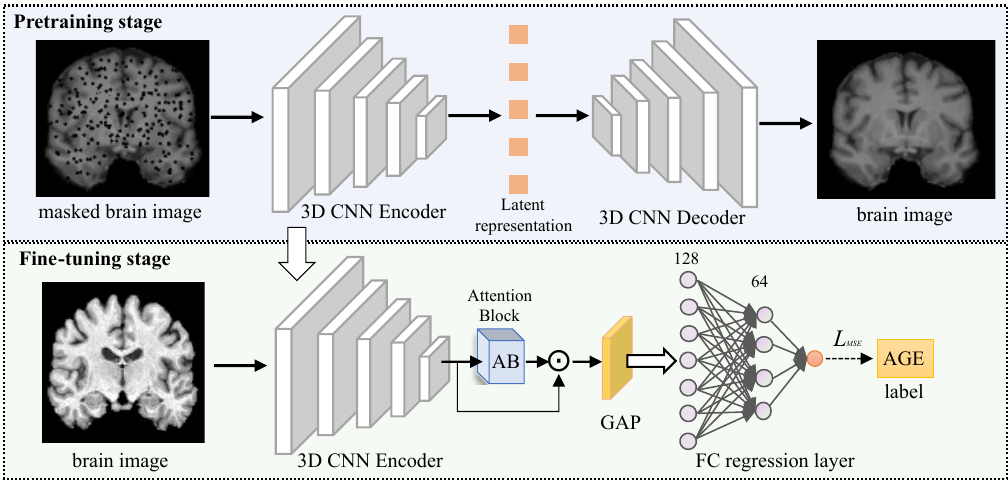}
\caption{Overview of the two-stage pretraining and fine-tuning procedure for brain age prediction.}
\label{fig:framework}
\end{figure*}

To systematically investigate this issue, this study employs the large-scale FOMO~\cite{munk2025large} brain MRI dataset as a representative example to examine how data characteristics influence self-supervised pretraining and downstream performance. Specifically, we perform self-supervised pretraining using a 3D Masked Autoencoder (MAE). The pretrained encoder is subsequently transferred and fine-tuned on our in-house labeled dataset for brain age prediction. The main contributions of this work including: (1) Systematic investigation of data quality factors contributing to brain MRI self-supervised pretraining performance. (2) Empirical validation through the brain age prediction task. (3) Analysis and visionary outlook on the evolving paradigms of brain imaging.

\section{Methodology}

Our task follows a two-stage framework, as illustrated in Fig.1. 
(1) A \textbf{self-supervised pretraining stage} based on a 3D Masked Autoencoder (MAE) that learns anatomical representations from unlabeled brain MRI scans; 
(2) A \textbf{downstream fine-tuning stage} for subject-level brain age prediction. 
This framework is designed to systematically evaluate how pretraining data quality affects downstream task performance.

\begin{figure*}[!t]
\centering
\setlength{\abovecaptionskip}{1pt} 
\includegraphics[width=0.97\linewidth]{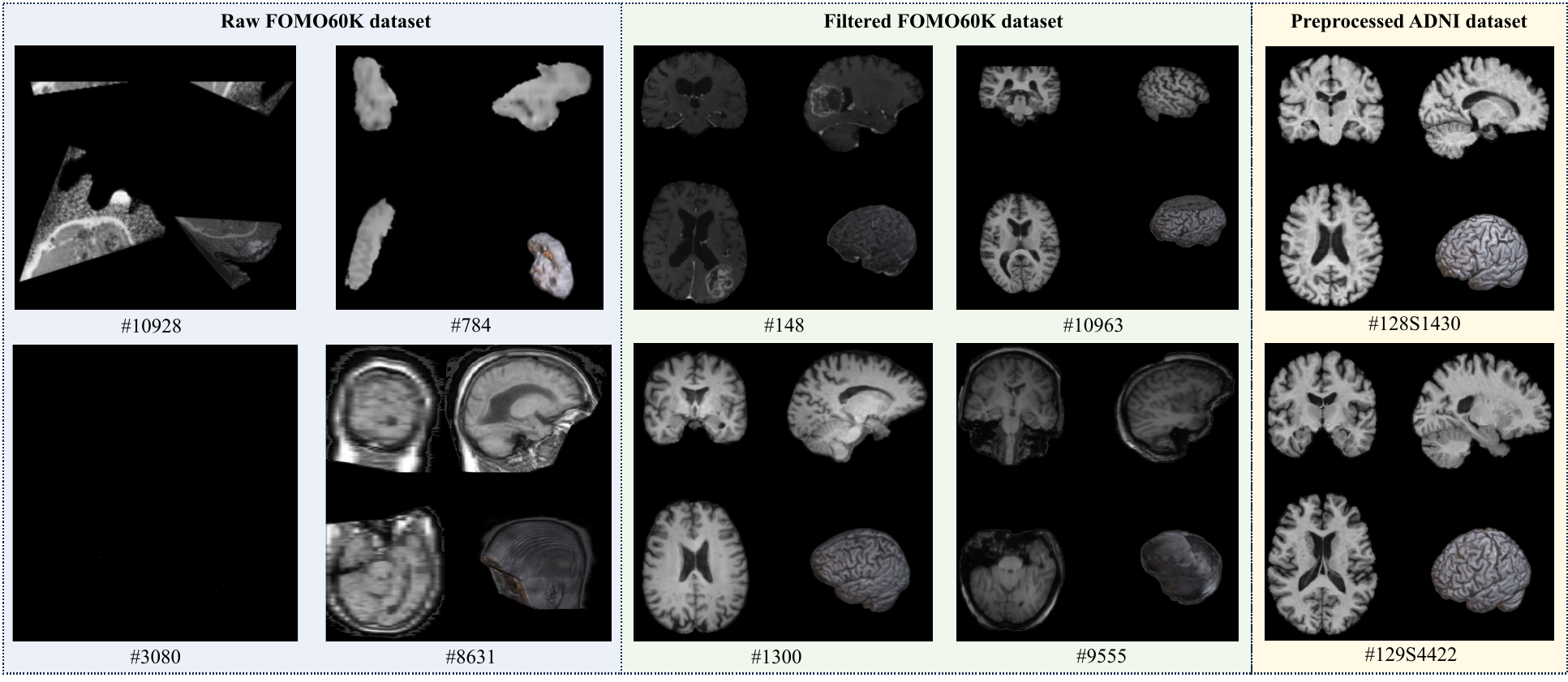}
\caption{Several representative brain MRI samples intentionally selected to reflect quality differences across the three pretraining datasets used in this study; each sample is labeled with its original data index for reference and review.}
\label{fig:samples}
\end{figure*}

\subsection{MAE Pretraining}

We adopt a 3D MAE architecture composed of a convolutional encoder and decoder, which effectively captures local anatomical continuity while maintaining computational efficiency for volumetric sMRI data. 
The encoder consists of sequential 3D convolutional blocks with channel dimensions of 16, 16, 32, 32, 64, 64, 128, and 128. Each block includes a 3×3×3 convolution, followed by 3D batch normalization and ReLU activation. During pretraining, a voxel-level random masking strategy is applied, where 50\% of voxels in each MRI scan are randomly masked, and only the remaining visible voxels are provided to the decoder for reconstruction.

The decoder takes the latent features generated by the encoder and reconstructs the entire MRI volume, including masked regions, through a symmetric 3D deconvolutional network. It contains four transposed convolutional layers with progressively reduced channel sizes (128→64→32→16→1). Each layer is followed by 3D batch normalization and ReLU activation, except for the final layer, which uses a sigmoid function to normalize voxel intensities to the range $[0,1]$. Finally, the reconstructed output is upsampled using trilinear interpolation to restore the original spatial resolution, allowing recovery of fine-grained anatomical details. The reconstruction objective is defined as the mean squared error (MSE) between the reconstructed and the input images.

\subsection{Fine-tuning for Brain Age Prediction}

After self-supervised pretraining, the encoder weights are transferred to a downstream CNN regression model for brain age prediction. The fine-tuning model retains the same CNN backbone as the pretrained encoder to ensure fair comparison. An attention block is introduced after the CNN encoder to highlight related brain regions~\cite{li2025transformer}, followed by a fully connected (FC) regression layer that produces the predicted age. 
The network is optimized using MSE loss between the predicted and ground-truth ages.

\begin{table*}[h]
\centering
\setlength{\abovecaptionskip}{3pt} 
\caption{Summary of downstream brain age prediction results. To clarify, the upper section (rows labeled as Raw and Filtered) corresponds to models pretrained leveraging the FOMO dataset, while the lower section presents the models pretrained leveraging the ADNI dataset. Model denotes the fine-tuning network (CNN is ACNN without the attention block); Data source specifies the pretraining dataset (‘–’ indicates no pretraining stage); Quantity represents the number of samples in the pretraining dataset; Image size is the resolution of MRI scans after resizing operation in pretraining, due to lots of raw brain scans in FOMO have one or more spatial dimensions with a resolution smaller than 20 voxels, it is infeasible to perform 3D CNN pretraining or using center-cropped (CC) operation; Frozen specifies whether encoder parameters were frozen during fine-tuning. In each section, the top two results are highlighted in bold to provide a clearer overall comparison.}
\scalebox{0.9}{
\begin{tabular}{l|c|c|c|c|ccc|ccc}
\cline{1-11}
\multirow{2}{*}{Model} & 
\multirow{2}{*}{Data source} & \multirow{2}{*}{Quantity} & \multirow{2}{*}{Image size}&
\multirow{2}{*}{Frozen} & \multicolumn{3}{c|}{Dataset I} & \multicolumn{3}{c}{Dataset II} \\ &&&& 
&CC ↑ &RMSE ↓  &MAE(yrs) ↓ &CC ↑ &RMSE ↓ &MAE(yrs) ↓ \\ \cline{1-11} 
ACNN  & Raw & 8017  & 121×145×121   & $\sqrt{}$  & 0.467 & 6.824  & 4.847  & 0.621  & 5.551 & 4.385  \\
ACNN   & Raw & 8017  & 204×240×150  & $\sqrt{}$  & 0.547 & 5.879 & 4.582  & 0.545 & 5.852 & 4.542  \\
ACNN   & Raw & 8017  & 204×240×150  & ×  & 0.575 & 5.987 & 4.685  & 0.663 & 5.564 & 4.364  \\
ACNN   & Filtered & 3511 & 204×240×150  & $\sqrt{}$   & 0.515 & 6.077 & 4.610   & 0.570 & 5.767 &  4.645   \\
ACNN   & Filtered  & 3511 & 204×240×150   &  ×  & 0.656 & 5.341 & \textbf{4.144}   &  0.630 & \textbf{5.253} & 4.251   \\
ACNN   & Filtered  & 3511 & CC204×240×150   & $\sqrt{}$  & 0.527 & 6.182 & 4.671  & 0.582 & 5.713 & 4.489  \\
ACNN   &  Filtered & 3511 & CC204×240×150  & × & \textbf{0.671} & \textbf{5.282} & 4.249  & \textbf{0.703} & \textbf{4.996} & \textbf{4.028}  \\
ACNN   & Filtered & 3511  & Original  & $\sqrt{}$   & 0.643 & 5.473 & 4.251  & 0.577 & 5.725 & 4.428  \\
ACNN   & Filtered & 3511  & Original  & ×   & \textbf{0.659} & \textbf{5.433} & \textbf{4.060}  & \textbf{0.675} & 5.376 & \textbf{4.155} \\ \hline
CNN   & – & –  & –  & –  & 0.605 & 5.815 & 4.704   &  0.629 & 5.417 & 4.181   \\
ACNN  & – & –  & –   &  –  & 0.534 & 6.004 & 4.450  & \textbf{0.671} & \textbf{5.060} & \textbf{3.961}  \\
ACNN   &  MRI  & 319 & 121×145×121 & $\sqrt{}$  & 0.595 & 5.681 & 4.370  &  0.575 & 5.714 & 4.552  \\
ACNN   &  MRI  & 319  & 121×145×121  & ×   & 0.656 & 5.349 & 4.189  & 0.618 & 5.648 & 4.405  \\
ACNN   &  PET  & 398 & 121×145×121  & $\sqrt{}$  & 0.573 & 5.955 & 4.564  & 0.634 & 5.447 & 4.289  \\
ACNN   & PET  & 398  & 121×145×121  &  ×  & \textbf{0.681} & \textbf{5.181} & \textbf{4.089}  & \textbf{0.688} & \textbf{5.157} & 4.092  \\
ACNN   &  MRI+PET  & 717 & 121×145×121  & $\sqrt{}$ &0.603 & 5.764 & 4.471  & 0.603 & 5.570 & 4.330  \\
ACNN   &  MRI+PET  & 717 & 121×145×121  &  ×  & \textbf{0.704} & \textbf{4.998}  & \textbf{3.923}  & 0.659 & 5.196 & \textbf{4.081} \\ 
\cline{1-11}
\end{tabular}
}
\end{table*}

\section{Experimental Results}
To systematically evaluate the impact and effect of data quality on pretraining performance, we first conducted pretraining on data sources of varying quality, followed by fine-tuning on a controlled ADNI~\cite{jack2015magnetic} dataset for the brain age prediction task. Finally, cross-comparisons and analyses were performed across multiple performance metrics.

\subsection{Dataset}
For the downstream brain age evaluation, we obtained a well-controlled dataset from  Alzheimer’s Disease Neuroimaging Initiative (ADNI). This data set includes 1,115 T1 sMRI scans from baseline, each scan corresponds to a distinct subject with a matched age label. All images were preprocessed following the standard pipeline~\cite{routier2021clinica}\cite{35}. The MRI scan has a resolution of 121×145×121. We divided the data into two distinct datasets, Dataset I and Dataset II, based on subject IDs.

In the pretraining stage, we trained MAE on three different unlabeled MRI datasets, some samples illustrated in Fig.2:

\begin{itemize}
    \item \textbf{Raw FOMO dataset:} The FOMO dataset is a large-scale and heterogeneous brain MRI dataset, where the scans of 13,900 sessions and 11,187 subjects, aggregated from 16 publicly available sources. FOMO exhibits a wide range of image resolutions and contrasts. Based on this dataset, we collected all FOMO T1 MRI 8017 scans for pretraining. Notably, a considerable portion of the brain scans show evident heterogeneity and artifacts, such as spatial distortion, missing tissues, and non-standard orientations.
    \item \textbf{Filtered FOMO dataset:} 
    This subset includes 3511 scans, which was obtained by removing scans with severe artifacts or imaging failures according to imaging quality and resolution criteria based on the raw FOMO dataset. The majority filtered MRI scans exhibit spatial dimensions no smaller than 204×240×150.
    \item \textbf{Preprocessed ADNI dataset:} 
    An additional set of 319 MRI and 398 PET scans was collected from the same preprocessing pipeline as the ADNI downstream tasks but without subject overlap among the 1,115 individuals used in the fine-tuning age prediction stage.
\end{itemize}

\subsection{Settings}
All hyperparameters were kept identical across datasets within the same experimental stage.
During pretraining, the learning rate was set to 1e-4 for the first. The batch size was 3, and training was performed for 60 epochs, selecting the model checkpoint with the lowest reconstruction loss. For experiments preserving the original image resolution, the batch size was set to 1, since the varying spatial dimensions of unresized MRI volumes prevented consistent batch alignment after encoding. During fine-tuning, the network was trained for 40 epochs with a learning rate of 1e-4.

\subsection{Comparative analysis and ablation.}
Table 1 summarizes the age prediction results of models fine-tuned after pretraining on different datasets. These results provide a comparative analysis of how the scale and quality of pretraining data influence downstream performance.

In brain age prediction experiments, we first compared the effects of multiple pretraining strategies. Under the same data split and hyperparameter settings, a model trained directly on the ADNI dataset was used as the baseline. The results showed that this model achieved better predictive performance than the models pretrained on the FOMO dataset and then fine-tuned. Specifically, pretraining on the raw FOMO data, on a filtered FOMO subset, or on variants whose MRI volumes were spatially resized, most failed to produce performance improvements in the downstream age prediction task.

To further investigate the impact of data quality on the effectiveness of pretraining, we conducted self-supervised pretraining on the preprocessed ADNI data set and fine-tuned the model for the brain age prediction. When using only sMRI data, performance did not improve, likely because the number of available samples was much smaller than that of FOMO, which restricts the pretraining stage from contributing meaningful representational gains. After adding PET data to form a multimodal pretraining set, additional accuracy gains were achieved. This suggests that richer and more diverse multimodal data help the model learn physiologically meaningful and cross-modality consistent representations, thereby enhancing generalization in downstream tasks. Nevertheless, even with PET added, the total sample size remained much smaller than the FOMO, indicating that in brain image pretraining, data quality and distributional alignment—rather than data set scale alone—are the key determinants of model performance.

\section{Prospect}
The effectiveness of brain image pretraining shows a complex and nonlinear dependence on data quality. Large-scale datasets alone do not ensure improved performance; low-quality samples can destabilize feature distributions and weaken model transferability. Thus, effective pretraining also depends on structural integrity and signal reliability.

Neuroimaging fundamentally differs from natural images in both physical acquisition and informational content. Brain MRI captures subtle anatomical and tissue variations rather than semantic cues such as color or texture, making learned features highly sensitive to artifacts and misregistration and limiting biological interpretability. Given the brain’s structural complexity, weak inter-subject differences are difficult to compare without preprocessing. Unlike natural scenes, brain images lack explicit semantic patterns, and even human observers struggle to visually assess task-relevant features. Inconsistent preprocessing further amplifies data heterogeneity, causing distribution shifts that undermine structural consistency and representation generalization. Together, these factors highlight the inherent complexity and sensitivity of self-supervised learning in neuroimaging. Therefore, directly transferring techniques from the computer vision domain to brain imaging tasks is not necessarily effective.

\section{Conclusion}

Taking brain age prediction as a representative task, this study systematically investigates the impact of data quality on large-scale self-supervised pretraining for brain imaging. Experimental results show that simply aggregating heterogeneous scans of varying quality does not effectively improve model performance. These findings highlight the critical role of data quality in representation learning, underscoring the need to build well-curated domain-specific datasets. They also reflect the unique challenges and complexities faced by computer vision methods when applied to clinical neuroimaging.

\section{Compliance with ethical standards}
\label{sec:ethics}
This research study was conducted retrospectively using human subject data made available in open access by ADNI and FOMO. Ethical approval was not required as confirmed by the license attached with the open access data.

\section{Acknowledgments}
\label{sec:acknowledgments}
Data used in preparation of this article were obtained from the Alzheimer's Disease Neuroimaging Initiative (ADNI) database. More details can be found at adni.loni.usc.edu. 
Data used in model pretraining were obtained from the large-scale FOMO brain MRI dataset, which can be found at huggingface.co/datasets/FOMO25/FOMO-MRI.

\bibliographystyle{IEEEbib}
\bibliography{refs}

\end{document}